\documentclass[10pt,twocolumn,letterpaper]{article}

\usepackage{iccv}
\usepackage{times}
\usepackage{epsfig}
\usepackage{graphicx}
\usepackage{amsmath}
\usepackage{amssymb}
\usepackage{enumitem}
\usepackage{subcaption}
\usepackage{amsthm}
\usepackage{multirow}
\usepackage[T1]{fontenc}

\usepackage[breaklinks=true,bookmarks=false]{hyperref}

\iccvfinalcopy 

\def\iccvPaperID{3632} 

\DeclareMathOperator{\EX}{\mathbb{E}}
\newtheorem{lemma}{Lemma}

\ificcvfinal\fi

\begin{document}

\title{Low Curvature Activations Reduce Overfitting in Adversarial Training}

\author{Vasu Singla \hspace{5mm} Sahil Singla \hspace{5mm}  Soheil Feizi \hspace{5mm} David Jacobs\\
University of Maryland\\
{\tt\small \{vsingla, ssingla, sfeizi, djacobs\}@cs.umd.edu}
}

\maketitle
\ificcvfinal\thispagestyle{empty}\fi

\begin{abstract}
    Adversarial training is one of the most effective defenses against adversarial attacks. Previous works suggest that overfitting is a dominant phenomenon in adversarial training leading to a large generalization gap between test and train accuracy in neural networks. In this work, we show that the observed generalization gap is closely related to the choice of the activation function. In particular, we show that using activation functions with low (exact or approximate) curvature values has a regularization effect that significantly reduces both the standard and robust generalization gaps in adversarial training. We observe this effect for {\it both} differentiable/smooth activations such as SiLU as well as non-differentiable/non-smooth activations such as LeakyReLU. In the latter case, the ``approximate'' curvature of the activation is low. Finally, we show that for activation functions with low curvature, the double descent phenomenon for adversarially trained models does not occur.
\end{abstract}

\section{Introduction}

Deep Neural Networks can be readily fooled by adversarial examples, which are computed by imposing small perturbations on clean inputs \cite{szegedy2014intriguing}. Adversarial attacks have been well studied in the machine learning  community in recent years \cite{carlini2017towards, madry2018towards, goodfellow2015explaining, papernot2016transferability, ge2021shift, laidlaw2019functional, laidlaw2021perceptual}.  There have been several defenses proposed against adversarial attacks in the literature \cite{papernot2016distillation, song2018pixeldefend, buckman2018thermometer}.
In our work we focus on adversarial training \cite{madry2018towards, goodfellow2015explaining, kurakin2017adversarial}, one of the most effective empirical defenses.


 Adversarial training involves training the network on adversarially perturbed data instead of clean data to produce a classifier with better robustness on the test set. However, it has been shown that networks produced through vanilla adversarial training do not robustly generalize well  \cite{schmidt2018adversarially, rice2020overfitting, farnia2018generalizable}. The gap between robust train and test accuracy for adversarially trained neural networks i.e. the \emph{robust generalization gap} can be far greater than the  generalization gap achieved during standard empirical risk minimization. In this work, we show that the robust generalization gap is significantly impacted by the curvature of the activation function, and activations with low curvature can act as efficient regularizers for adversarial training, effectively mitigating this phenomenon. 

 \begin{figure}[t]
\begin{center}
\includegraphics[width=0.45\textwidth]{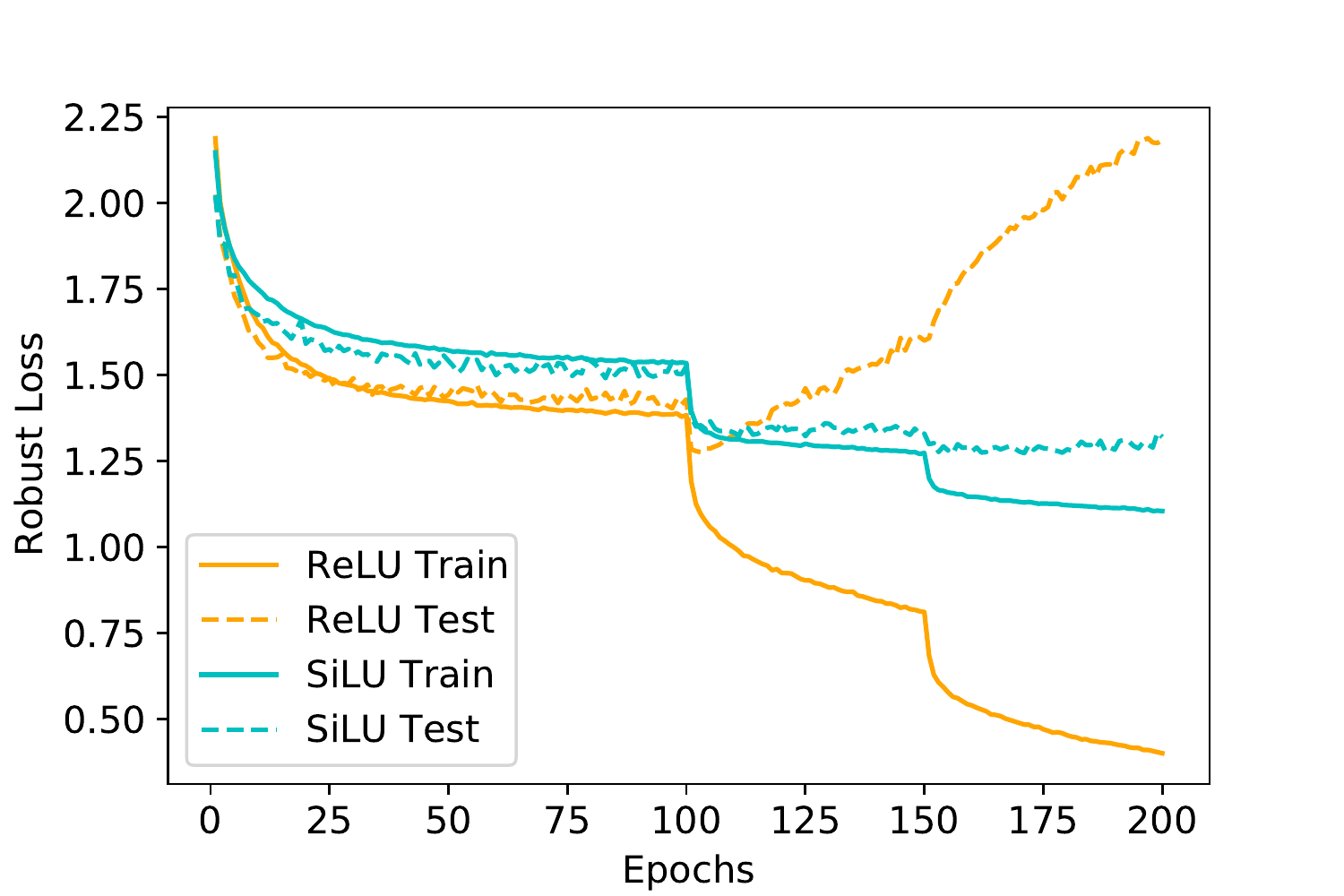}
\end{center}
\vspace{-2mm}
\caption{Learning curves for a robustly trained ResNet-18 model on CIFAR-10. Using an activation function with low curvature such as SiLU prevents robust overfitting, achieving and maintaining low test robust loss, even compared to the best early-stop checkpoint of a network with ReLU activation function. The learning rate is decreased by a factor of 10 at the 100th and 150th epoch.}
\vspace{-5mm}
\label{fig:relu_vs_swish}
\end{figure}

  Rice \etal \cite{rice2020overfitting} showed for adversarially trained ReLU networks, the best robust test accuracy is not achieved by allowing models to train until convergence. Adversarial training has the characteristic that, after a certain point, further training will continue to decrease the robust training loss, while the robust test loss starts increasing. This phenomenon is referred to as \emph{robust overfitting} and ultimately leads to poor robust accuracy on the test set.
 Rice \etal also showed that while traditional approaches against overfitting such as $l_1$, $l_2$ regularization can mitigate robust overfitting, no approach works better than simple early stopping. Since standard accuracy continues to improve even after the network overfits to adversarial examples, early stopping leads to trade-off between selecting a model with high robust accuracy versus a model with high standard accuracy \cite{chen2020adversarial}. 

 In this work, we systematically study the impact of activation functions on generalization. We first theoretically analyze the relation between maximum curvature of the activation function and adversarial robustness. A key observation of our paper is that for smooth activation functions the maximum value of the second derivative of the function, i.e. the maximum curvature has a significant impact on robust generalization. Specifically by using activations with low curvature the robust generalization gap can be reduced, whereas with high curvature the robust generalization gap increases.  For instance, in Figure 1 for an adversarially trained CIFAR-10 model, test error on adversarial examples for the ReLU activation function decreases after the first learning rate drop, and keeps increasing afterwards. However, for SiLU \cite{ramachandran2017searching} a smooth activation function with low curvature, robust test loss keeps decreasing. We also show that the choice of activation has a similar effect on the standard generalization gap. In other words, activations that show a large robust generalization gap also have a large standard generalization gap,  and vice versa. Our work therefore provides novel insights to the robust overfitting phenomenon. The main objective of our work is to understand the relation between curvature of the activation function and adversarial training, and highlight findings which can be useful for training adversarially robust models.

 Xie \etal \cite{xie2020smooth} showed that replacing ReLU, a widely used activation function, by ``smooth" \footnote{We use the same definition of smoothness as Xie \etal, that the function is $C^1$ smooth, that is, that the first derivative is continuous everywhere.} activation functions such as Softplus or SiLU with a weak adversary (single step PGD), improves adversarial robustness on Imagenet \cite{imagenet_cvpr09} for ``free". They posit smooth activations allow adversarial training to find harder adversarial examples and compute better gradient updates to weight parameters. Further works have however  demonstrated that while smooth activation functions can positively affect clean and robust accuracy, the trend is not as clear as the one observed by Xie \etal. Thus, ReLU networks remains a prominent choice for robust classification \cite{gowal2020uncovering, pang2021bag}.
 
 In contrast to Xie \etal \cite{xie2020smooth}, we consider a strong adversary for training and show that smoothness of activations is not required to obtain a regularization effect on adversarial training. In our experiments, we show that the same regularization can be achieved using non-smooth activations with low ``approximate"" curvature. For non-smooth activations however, curvature is not well-defined. We consider LeakyReLU which is a non-smooth activation function and use the difference of activation slopes in positive and negative domains as the approximate maximum curvature of the activation function. Even for such a non-smooth activation function, we observe that if the approximate curvature is low, the robust overfitting phenomenon does not occur. Also in contrast to Xie \etal \cite{xie2020smooth} we empirically show that smooth activations can perform worse than ReLU, if the smooth activation has high curvature. 


Finally, we study the phenomenon of double descent generalization curves seen in standard training \cite{belkin2019reconciling} and  robust training \cite{nakkiran2020deep}. Double descent describes the following phenomenon. 
Increasing model complexity causes test accuracy to first increase and then decrease. Then upon reaching a critical point known as interpolation threshold, test accuracy starts increasing again.
We show that double descent curves reported by \cite{rice2020overfitting} for robust overfitting using ReLU do not hold for activation functions with low curvature such as SiLU. 

\begin{figure*}[t]
\begin{center}
\includegraphics[scale=0.35]{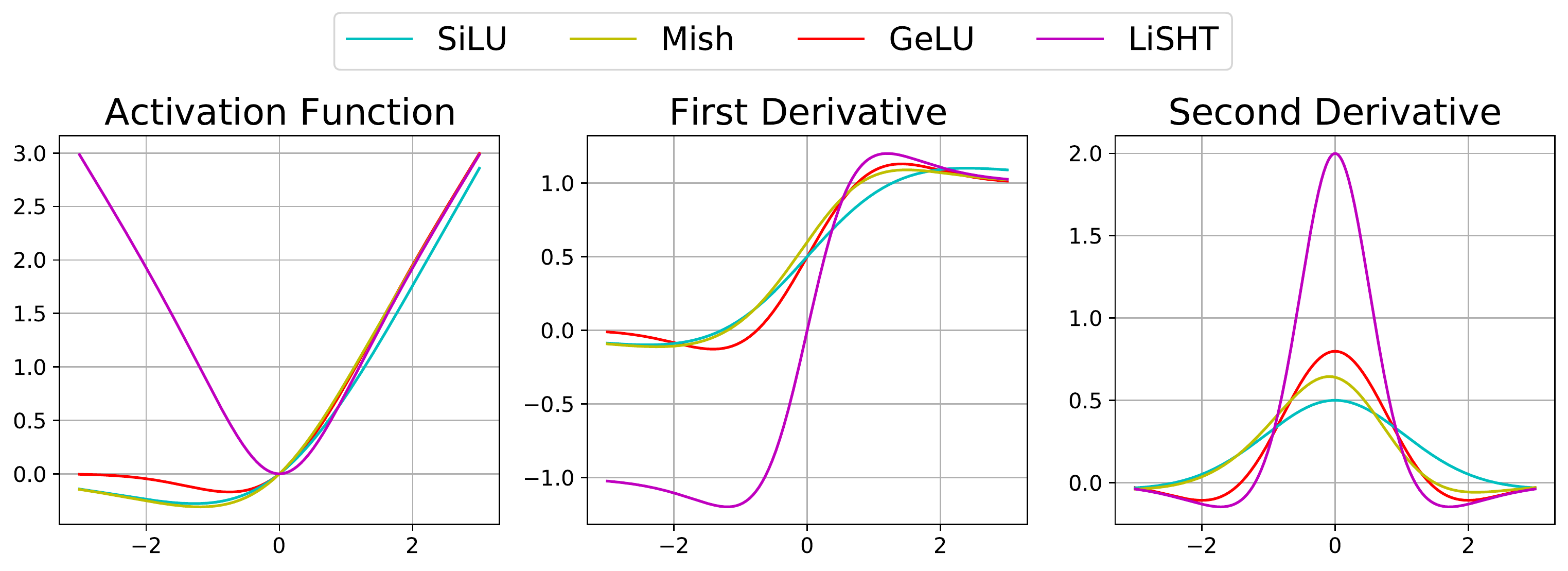}
\end{center}
\vspace{-2mm}
\caption{Activation functions along with their first and second derivatives.}
\label{fig:smooth_activations}
\vspace{-3mm}
\end{figure*}

\section{Related Works}

Goodfellow \etal \cite{goodfellow2015explaining} provided one of the first approaches for adversarial training based on generating adversarial examples through the fast sign gradient method (FGSM). Building on this, a stronger adversary known as the basic iterative method \cite{kurakin2017adversarial} was proposed in subsequent work, using multiple smaller steps for generating adversarial examples. Madry \etal \cite{madry2018towards} extended this adversary with multiple random restarts to train models on adversarial data, referred to as projected gradient descent (PGD) adversarial training. Further works have focused on improving the performance of the adversarial training procedure with methods such as feature denoising \cite{xie2019feature}, hypersphere embedding \cite{pang2020boosting}, balancing standard and robust error \cite{zhang2019theoretically} and using friendly adversarial data \cite{zhang2020attacks}. A separate line of works has focused on speeding up adversarial training due to its increased time complexity, by reducing attack iterations and computational complexity for calculating gradients \cite{zhang2019you, shafahi2019adversarial, wong2020fast}. Another tangent line of work focuses on adversarial training for universal attacks \cite{shafahi2020universal, benz2021universal}.

Besides adversarial training, several other defenses have been proposed such as defensive distillation \cite{papernot2016distillation}, preprocessing techniques \cite{guo2018countering, song2018pixeldefend, buckman2018thermometer} and randomized transformations \cite{xie2017mitigating, dhillon2018stochastic, liu2018towards} or detection of adversarial examples \cite{metzen2017detecting, feinman2017detecting}.  However these methods were later broken by stronger adversaries \cite{athalye2018obfuscated, tramer2020adaptive, carlini2017adversarial}. These defense methods were shown to rely on obfuscated gradients (gradient masking), which provided a false sense of security. Due to the bitter history of gradient masking as a defense, Xie \etal \cite{xie2020smooth} proposed use of smooth activations with a single step PGD attack, reaching state of the art robust performance on ImageNet \cite{imagenet_cvpr09}. Xie \etal hypothesize that using smooth activations provides networks with better gradient updates and allows adversaries to find harder examples. 

Since many defenses proposed in the literature have been broken, another separate line of work has focused on certified defenses, which can guarantee robustness against adversarial attacks.  These methods use techniques such as mixed-integer programming methods \cite{tjeng2018evaluating, lomuscio2017approach, fischetti2017deep, bunel2018unified} and satisfiability modulo theories \cite{katz2017reluplex, ehlers2017formal, huang2017safety}. Some certification methods bound the global Lipschitz constant of the network, which are usually loose for large neural networks with multiple layers \cite{anil2019sorting, gouk2021regularisation}. Another line of work has focused on providing loose certificates using other techniques such as randomized smoothing \cite{cohen2019certified, lecuyer2019certified, levine2019certifiably, levine2020derandomized}, abstract representations \cite{gehr2018safety, mirman2018differentiable, singh2019abstract}, interval bound propagation, second-order information \cite{singla2020secondorder}, \cite{gowal2019scalable} and duality and linear programs \cite{salman2020convex, wong2018provable, wong2018scaling}.

Lack of overfitting in overparameterized deep learning models is an intriguing phenomenon for deep learning \cite{zhang2021understanding}. These models can be trained to effectively zero training error, without having impact on test time performance. Hence, it is now standard practice in deep learning to train longer and use large overparameterized models, since test accuracy generally improves past an interpolation point also known as double descent generalization \cite{belkin2019reconciling, nakkiran2020deep}. Schmidt \etal \cite{schmidt2018adversarially} however have shown that sample complexity required for adversarially robust generalization is significantly higher than sample complexity for standard generalization. In a recent work, Rice \etal \cite{rice2020overfitting} have shown the overfitting phenomenon to be dominant in adversarial training and show that training longer decrease robustness on test data. Rice \etal also show that double descent generalization curves seem to hold with increase in model size but not by training longer. A recent work shows that robust overfitting may be mitigated \cite{chen2021robust} using a combination of previously proposed techniques such as knowledge-distillation \cite{yuan2020revisiting} and stochastic weight averaging \cite{izmailov2018averaging}. Another recent work, proposes the use of adversarial weight perturbations \cite{wu2020adversarial} to mitigate robust overfitting, which may also increase the training time. AVMixup \cite{lee2020adversarial} also discussed the idea of robust overfitting and proposed a combination of AVMixup, Label smoothing and  Feature Scatter to alleviate robust overfitting on CIFAR-10. In contrast to these works, we discover a novel way to mitigate this phenomenon without using complex regularization techniques that may lead to additional hyper-parameters and increased training time; we only modify the activation function of the network.  

\section{Background}

\subsection{Adversarial Training}
To train networks that are robust to adversarial examples, the following robust optimization framework is used:
$$
    \min_{w} \EX_{(x, y) \sim \mathbb{D}} \left[ \max_{d(x, \hat{x}) \leq \epsilon} l(f_{w}(\hat{x}), y) \right]
$$
where $x$ is a training sample with ground truth label $y$ sampled from the underlying data distribution $\mathbb{D}$,  $l(., .)$ is the loss function, $f_{w}$ is the model parameterized by $w$ parameters, $d(., .)$ is a distance function and $\epsilon$ is the maximum distance allowed.  Typically, the distance function is chosen to be an $l_p$-norm ball such as the $l_2$ and $l_{\infty}$-norm balls though other non $l_p$ threat models have been considered in \cite{laidlaw2019functional, laidlaw2021perceptual}. Adversarial training thus consists of two optimization problems, the inner maximization problem to construct adversarial samples, and the outer minimization problem to update weight parameters $w$. To solve the inner maximization problem, different types of attacks have been used in the literature, such as projected gradient descent (PGD) \cite{madry2018towards} or fast gradient sign method (FSGM) \cite{wong2020fast}. 
For example, an $l_{\infty}$ PGD adversary starts with a random initial perturbation drawn from a uniform distribution $\mathcal{U}$, and iteratively adjusts the perturbation with $\alpha$ step-size towards the $l_{\infty}$ gradient direction, followed by projection back onto the $l_{\infty}$ norm ball with maximum radius $\epsilon$:
\begin{align*}
\hat{x}_{0} &= x + \mathcal{U}(-\epsilon, \epsilon) \\
\bar{x}_t &= \hat{x}_{t} + \alpha \cdot \text{sign} \big( \nabla_{\hat{x}_{t}} l(f(\hat{x_{t}}), y) \big) \\
\hat{x}_{t+1} &= \max(\min(\bar{x}_t ,x + \epsilon), x - \epsilon) \end{align*} 
\begin{figure}[t]
\centering
\includegraphics[scale=0.55]{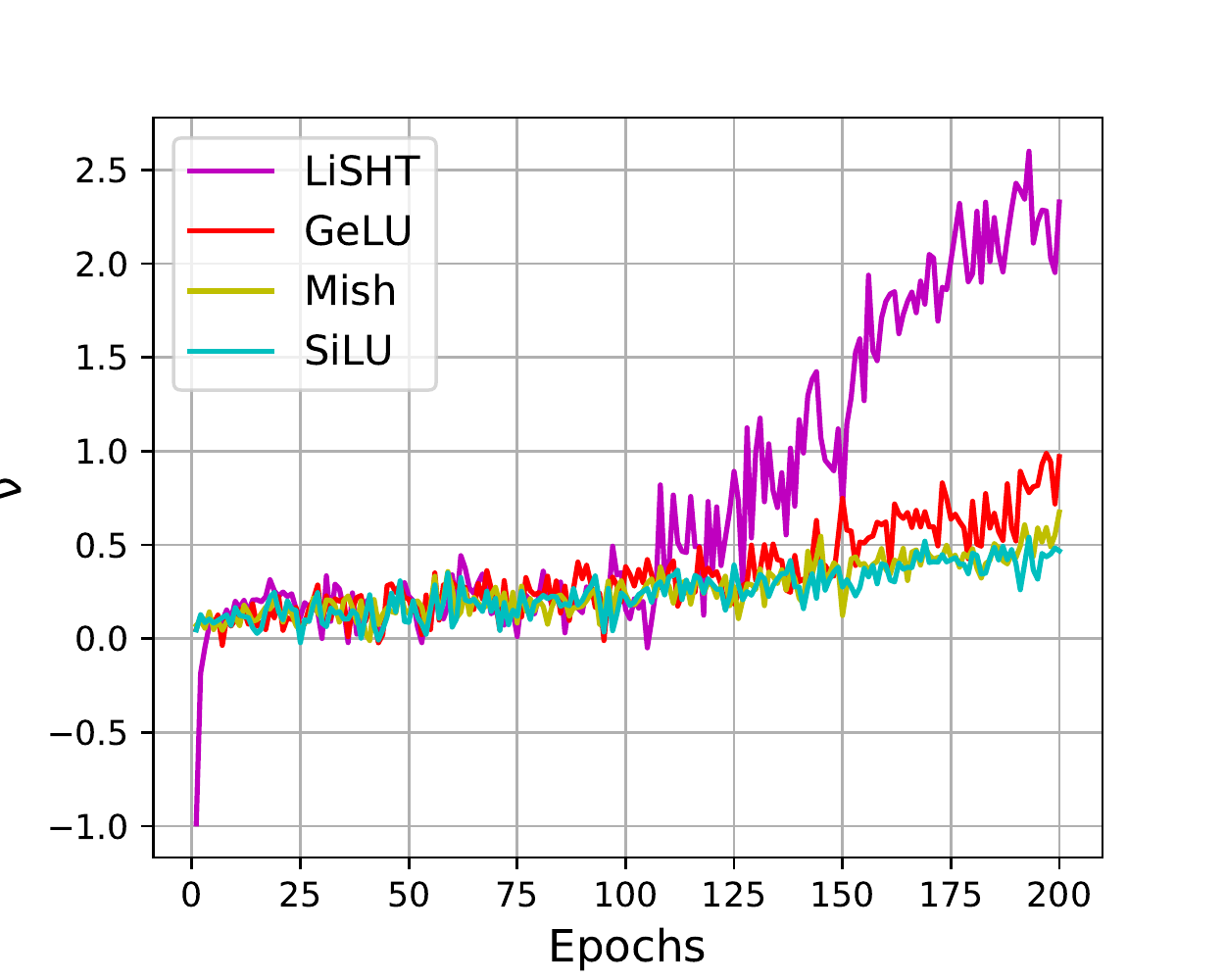}
\vspace{-2mm}
\caption{Maximum eigenvalues for a batch of test examples for Resnet-18 models with different smooth activations. Eigenvalues are larger for activations with high curvature.}
\label{fig:hessian}
\vspace{-6mm}
\end{figure}

\subsection{Robust Overfitting}

A surprising characteristic of overparametrized models is their good generalization behavior observed in practice.   \cite{belkin2019reconciling, neyshabur2017exploring}. Although overparameterized models have enough model complexity to memorize the dataset even on random labels \cite{zhang2021understanding}, they can be trained to zero error on the training set with no detrimental effects on generalization. For the standard (non-adversarial) empirical risk minimization setting, modern convergence curves indicate that while training for long periods of time, test loss continues to decrease  \cite{nakkiran2020deep}. This has led to the practice of training models for as long as possible to achieve better generalization \cite{hoffer2017train}. However, in adversarial training it was noted that training longer can cause overfitting and result in worse test performance \cite{rice2020overfitting}. This phenomenon has been referred to as "robust overfitting" and shown to occur with a variety of datasets, model architectures and different threat models. 

Regularizers are standard tools in practice to mitigate the effects of overfitting, especially in the regime when the number of parameters are larger than the number of data points. The standard regularization techniques such as $l_1$ and $l_2$ regularization and data augmentation methods such as Cutout \cite{devries2017cutout} and Mixup \cite{zhang2018mixup} have been shown to be ineffective against robust overfitting phenomena \cite{rice2020overfitting}.  Only early-stopping using a validation dataset and semi-supervised learning methods that augment the dataset with unlabelled data have been shown to be effective and reduce the generalization gap for adversarially robust learning. Data augmentation using semi-supervised methods however requires the use of additional data that may not be available. Early stopping leads to selection of an earlier checkpoint and causes a trade off between robust accuracy and standard accuracy, as training longer leads to better standard test accuracy.


\section{Impact of Activation Curvature on Adversarial Training} \label{activation_main_section}

In this section we consider the effects of curvature for smooth activation functions on standard and robust generalization gaps. We define curvature for smooth activation functions by the maximum of the second derivative\footnote{Note that this definition of curvature is different from the standard definition of curvature used for twice-differentiable functions.} i.e $\max_{x} f''(x)$. We consider the following smooth activation functions, which are ranked by decreasing curvature as follows  (see Figure \ref{fig:smooth_activations} for functions and their first and second derivatives): 
\begin{enumerate}[leftmargin=*]
\item \textbf{Linearly Scaled Hyperbolic Tangent} (LiSHT) \cite{roy2019lisht}: $f(x) = x*\tanh(x)$, this function has highest curvature among activations considered.

\item \textbf{Gaussian Error Linear Unit} (GeLU) \cite{hendrycks2016gaussian}: $f(x) = x*\Phi(x)$, where $\Phi(x)$ is gaussian cummulative distribution function.

\item \textbf{Mish} \cite{misra2020mish}: $f(x) = x*\tanh(ln(1+\exp(x)))$ is a smooth continuous function similar to SiLU.

\item \textbf{SiLU} \cite{ramachandran2017searching}: $f(x) = x*sigmoid(x)$ is a smooth approximation to ReLU but has a non-monotonic ``bump" for $x < 0$.  
\end{enumerate}

We also conduct experiments for non-smooth ReLU activation as a baseline. Code for reproducing our experiments can be found at \text{https://github.com/vasusingla/low\_curvature\_activations}.

\begin{figure*}[tp]
  \begin{subfigure}[t]{0.35\textwidth}
    \includegraphics[scale=0.415]{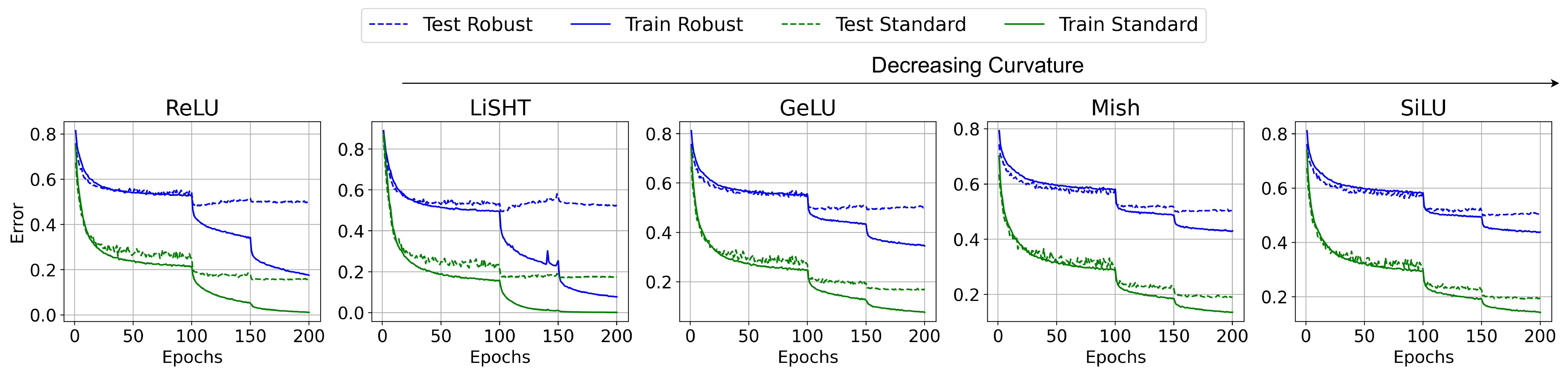}
    \label{fig:error_curves}
      \includegraphics[scale=0.35]{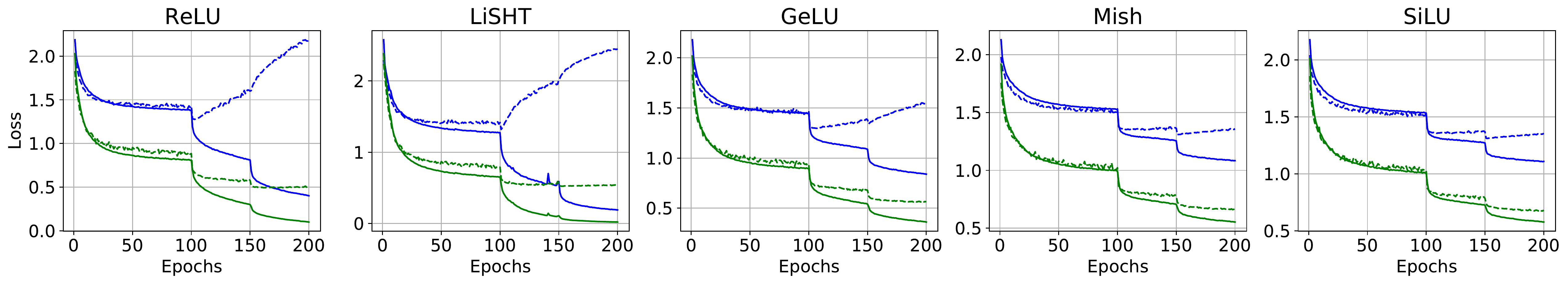}
    \label{fig:loss_curves}
  \end{subfigure}
\vspace{-2mm}
\caption{Learning curves for CIFAR-10 dataset on Resnet-18 for different activation functions. ReLU activation is non-smooth and included as a baseline, all the other activations are ordered by decreasing curvature from left to right. Top graphs show standard and robust error, and bottom graphs represent loss curves for both train and test data.}
\label{fig:training_curves}
\vspace{-3mm}
\end{figure*}

\subsection{Analyzing the influence of activations on robustness}
In this section, we analyze the theoretical relationship between curvature of the activation function and adversarial robustness. The motivation behind our analysis is to provide an intuition for our observations, we do not rigorously prove a monotonic relationship between robustness and activation curvature. To elucidate this, we first consider the relation between the input Hessian (ie., the second derivatives of the output with respect to the input) and adversarial robustness. 
We consider a simple binary classifier $f$, implemented as a two-layer neural network. Let $w_1, w_2$ be weight matrices for the first and second layers respectively. Let $\sigma(\cdot)$ be a twice differentiable activation function and $\sigma''(\cdot)$ denote the second derivative of the activation function. The two layer neural network can then be represented as $f(x) = w_2^T \sigma(w_1 x )$. Assume the final layer of the network outputs a single logit, which is transformed to probability using a sigmoid function. In other words, the probability of a sample being in class 0 is given as $p(x) = sigmoid(f(x))$. Assuming a sample is classified into class 1 if $p(x) < 0.5$, then a sample $x$ is classified into class 1 iff $f(x) < 0 $ and class 0 otherwise. In other words, we use a probability threshold of $0.5$, to classify an example into class 1.
 We assume that the neural network can be locally well approximated using the second order Taylor expansion. We now use the results by \cite{moosavi2019robustness} about the relation between the input Hessian and robustness.
Let $x$ belong to class 1, then for $x+\delta$ to be classified as class 0, the minimal $l_2$ perturbation that fools the classifier can be written as:
\begin{gather*}
\delta^{*} = \arg \min_{\delta} \|\delta\| \\
\text{s.t. } f(x) + \nabla_x f(x)^{T} \delta + \frac{1}{2} \delta^T \nabla_x^2 f(x) \delta \geq 0
\end{gather*}
It can be shown under these assumptions the magnitude of $\delta^*$ can be upper and lower bounded with respect to input curvature. We use the following lemma - 
\begin{lemma} \cite{moosavi2019robustness}
 Let $x$ be such that $c = -f(x) \geq 0$, and let $g = \nabla_{x} f(x)$. Assume that $\nu = \lambda_{max} \Big(\nabla_x^2 f(x) \Big) \geq 0$, denotes the largest eigenvalue and let $u$ be the eigenvector corresponding to $\nu$. Then,
\begin{equation}
    \begin{aligned}
\frac{\|g\|}{\nu} \Bigg(\sqrt{1 + \frac{2 \nu c}{\|g\|^2}}-1 \Bigg) &\leq \|\delta^*\| \\
&\leq  \frac{\|g^T u\|}{\nu} \Bigg( \sqrt{1 + \frac{2\nu c}{(g^Tu)^2}} - 1 \Bigg)
    \end{aligned}
\end{equation}
\end{lemma}
This lemma shows that upper and lower bounds on the magnitude of  $\delta^*$ increase as $\nu$ decreases keeping all other factors constant \cite{moosavi2019robustness}. An increase in $\|\delta^*\|$ therefore increases the minimum $l_2$ ball  required to find an adversarial example for input $x$, leading to increased robustness. Therefore, a low maximum eigenvalue of the input Hessian leads to higher adversarial robustness.


\begin{table*}[t]
 \centering
 \small
\begin{tabular}{|c|c|c|c|c|c|c|c|c|c|}
\hline
\multirow{2}{*}{Dataset} & \multirow{2}{*}{Activation} & \multicolumn{4}{c|}{Robust Accuracy} & \multicolumn{4}{c|}{Standard Accuracy} \\ \cline{3-10} 
 &  & Final Train & Final Test & Best Val & Diff. & Final Train & Final Test & Best Val & Diff. \\ \hline
\multirow{5}{*}{CIFAR-10} & LiSHT & 92.27 & 47.21 & 50.31 & \textbf{45.06} & 99.9 & 82.53 & 82.44 & \textbf{17.37} \\
 & ReLU & 82.46 & 49.25 & 51.06 & \textbf{33.21} & 98.9 & 83.73 & 81.62 & \textbf{15.17} \\
 & GeLU & 65.45 & 49.31 & 50.15 & \textbf{16.14} & 92.41 & 82.81 & 79.25 & \textbf{9.6} \\
 & Mish & 57 & 49.18 & 49.62 & \textbf{7.82} & 86.48 & 80.05 & 79.96 & \textbf{6.43} \\
 & SiLU & 56.15 & 48.91 & 49.41 & \textbf{7.24} & 85.79 & 80.55 & 80.57 & \textbf{5.24} \\ \hline
\multirow{5}{*}{CIFAR-100} & LiSHT & 93.58 & 18.62 & 22.48 & \textbf{74.96} & 99.92 & 49.12 & 49.13 & \textbf{50.8} \\
 & ReLU & 79.87 & 18.81 & 25.91 & \textbf{61.06} & 98.58 & 51.58 & 51.05 & \textbf{47} \\
 & GeLU & 57.96 & 21.56 & 26.33 & \textbf{36.4} & 89.18 & 53.67 & 49.5 & \textbf{35.51} \\
 & Mish & 39.65 & 24.27 & 25.88 & \textbf{15.38} & 71.5 & 53.43 & 48.37 & \textbf{18.07} \\
 & SiLU & 37.81 & 24.29 & 25.82 & \textbf{13.52} & 68.73 & 52.65 & 52.18 & \textbf{16.08} \\ \hline
\end{tabular}
\caption{Performance of different activations on CIFAR-10 and CIFAR-100 with ResNet-18. We use the best checkpoint based on \textbf{best robust accuracy} on the validation set shown in ``Best Val"" column.  The generalization gap, i.e difference between final train and final test accuracy is shown in ``Diff." column. Generalization gap for both standard and robust accuracy increases for activations with high curvature. }
\label{tab:robust_data}
\vspace{-3mm}
\end{table*}

We now show the relation between activation functions and input curvature. For the considered two layer neural network, the Hessian with respect to the input $x$ is given as:
\begin{equation}\label{eq:hessian}
    \nabla_{x}^{2} f(x) = w_1^{T} diag \Big( \sigma''(w_1 x) \odot  w_2 \Big) w_1 
\end{equation}
    where $\odot$ denotes the Hadamard product between two vectors. Equation \ref{eq:hessian} shows that the Hessian of the input directly depends on $\sigma''(.)$, which suggests that an increase in the curvature of the activation function leads to an increase in the norm of the input Hessian.   Finally, although we assume our activation to be smooth we expect similar results for non-smooth activations. 

 We empirically show the relation between $\nu$ and activation curvature holds for adversarially trained Resnet-18 models. The learning curves presented in Fig. \ref{fig:hessian} show that for activations with high curvature, the maximum eigenvalue of the input Hessian indeed is larger.  This result combined with our previous observation therefore suggests high activation curvature indeed leads to lower robustness.

\subsection{Activation Curvature and Generalization Gap} \label{activation_curvature}
In this section we show results for the adversarial training for different smooth activation functions. We hypothesize that for adversarial trained networks, activations with low curvature are more robust and have a small generalization gap. 

\textbf{Experimental Settings - } We show our results on the CIFAR-10 and CIFAR-100 dataset \cite{krizhevsky2009learning}. For comparison with best early-stop checkpoint \cite{rice2020overfitting}, we randomly split the original set into training and validation set with $90\%$ and $10\%$ of the images respectively. We consider the $l_\infty$ threat model and use PGD-10 step attack with a single restart for training and PGD-20 step attack with 5 restarts for reporting the test accuracy. For the attack hyper-parameters, we use $\epsilon=8/255$ and $\alpha=2/255$. We use the ResNet-18 \cite{he2016deep} architecture for all our experiments except for experiments with double descent curves where we use Wide ResNet-28 \cite{zagoruyko2016wide}. We use the same training setup as \cite{rice2020overfitting} throughout the paper, an SGD optimizer with momentum of $0.9$ and weight decay $5 \times 10^{-4}$ for 200 epochs with batch size of 128.

We discover that choice of activation function has a large impact on robust overfitting. Figure \ref{fig:training_curves} shows our results.  First we reproduce the effect of robust overfitting observed by Rice \etal \cite{rice2020overfitting} for all the activations. The robust training loss keeps decreasing, however robust test loss rises shortly after the first learning rate drop. For standard training and standard test loss however, both keep decreasing throughout training. Training appears to proceed smoothly at the start, however at the learning rate drop on the 100th and 150th epochs, robust test error decreases briefly and then keeps increasing as training progresses. This phenomenon shows the best performance for robust test accuracy is \textbf{not} achieved by training till convergence, unlike standard training. In contrast the best standard accuracy for adversarial training is still reached by training till convergence. We show that for activation functions with lower curvature the robust overfitting phenomenon occurs to a lesser degree. In contrast to Xie \etal \cite{xie2020smooth}, we also show that LiSHT a smooth activation function performs worse than the non-smooth ReLU function and shows a larger robust generalization gap as shown in Fig. \ref{fig:training_curves}. We also note that \emph{for activations that display a large robust generalization gap, the standard generalization gap is also higher}. Finally, the curvature of the activation function has a direct impact on both the robust and standard generalization gaps, as shown in the learning curves. For activations with high curvature such as LiSHT and GeLU the generalization gap is large and for activations with low curvature such as Mish and SiLU the generalization gap is much lower. Note that although the training loss/error is higher for activations with lower curvature, adversarial training is much more stable and allows training till convergence, achieving better standard accuracy and maintaining similar robust accuracy.

We show the quantitative results in Table \ref{tab:robust_data}.  To show the gap due to robust overfitting (decay in performance from peak robust accuracy) we also show the best robust accuracy found using early stopping with a validation set. We also report the corresponding standard accuracy for the \textbf{best robust accuracy checkpoint} (not the best standard accuracy checkpoint). The robust and standard generalization gap decrease for CIFAR-10 and CIFAR-100 as seen in Table \ref{tab:robust_data}. The effects of robust overfitting,  (i.e difference in best and final checkpoint on robust accuracy) also decreases for activations with smaller curvature. For example, the overfitting gap falls from $3.1\%$ for LiSHT to $0.5\%$ for SiLU on CIFAR-10. Standard accuracy however, either remains the same or improves by training longer (compared to the best checkpoint). On CIFAR-100 upon training till convergence, \emph{SiLU simultaneously achieves both robust and standard accuracy higher than ReLU.} These results therefore validate our claim that low curvature activations reduce robust overfitting. Using the best validation checkpoint for CIFAR-100, \emph{SiLU achieves nearly the same robust accuracy and higher standard accuracy than ReLU.} The results therefore show that for adversarial training, curvature of the activation function play an important role in obtaining high robust and standard accuracy.


\begin{figure}
     \centering
         \centering
         \includegraphics[width=0.32\textwidth]{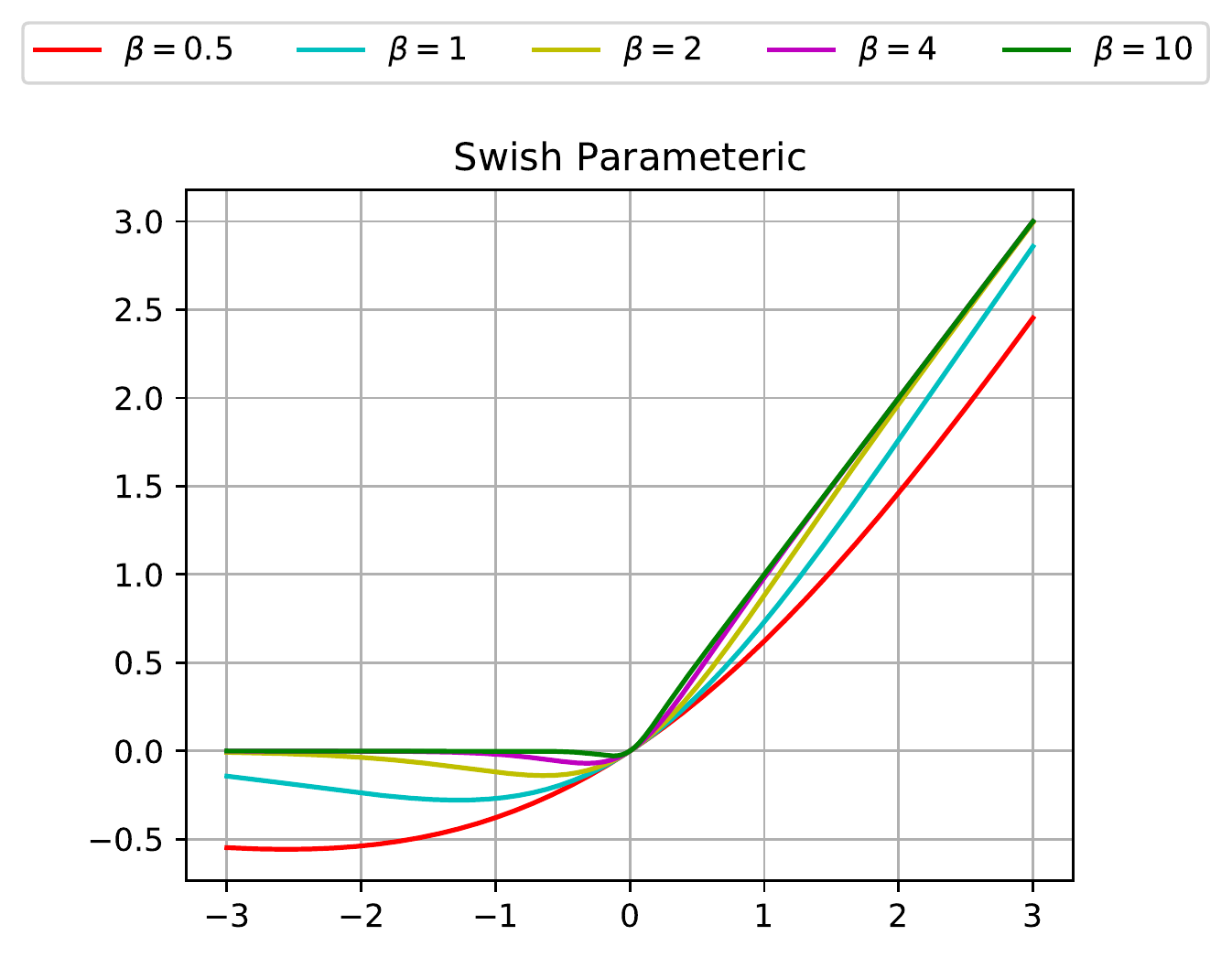}
         \caption{Visualization of PSwish with different $\beta$ values.}
         \label{fig:parameteric_swish}
\vspace{-2mm}
\end{figure}

\begin{table}[t]
    \centering
    \small 
    \begin{tabular}{c|c|c|c|c|c|c}
    $\beta$ & \multicolumn{3}{|c}{Robust Accuracy} & \multicolumn{3}{|c}{Standard Accuracy} \\
     & Train &  Test & Diff. & Train & Test  & Diff.  \\
    \cline{1-7}
    0.5 & 47.00 & 45.24 & \textbf{1.76} & 75.39 & 73.57 & \textbf{1.82} \\
    1  & 56.15 & 48.91 & \textbf{7.31} & 85.79 & 80.55 & \textbf{5.24}\\
    2 & 69.65 & 49.6 &  \textbf{20.05} & 94.57 & 83.39 & \textbf{11.18}\\ 
    4  & 83 & 49.92 & \textbf{33.08} & 98.82 & 84.48 & \textbf{14.34} \\
    10 & 89.2 & 50.63 & \textbf{38.57} & 99.7 & 83.57 & \textbf{16.13} \\
    \end{tabular}
    \caption{Performance of PSwish with different $\beta$ values, higher $\beta$ value indicates higher curvature. Results are shown for final checkpoint and show that for activations with high curvature, standard and robust generalization gap increases. }
    \label{tab:parameteric_swish}
    \vspace{-3mm}
\end{table}

\subsection{Curvature effects with Parameteric Swish} 
To further understand the impact of activation curvature on standard and robust generalization gap, we conduct analysis with \emph{Parameteric Swish (PSwish)} \cite{bingham2021discovering}, defined as follows:
$$
f(x) = x \cdot sigmoid(\beta x)
 $$
The SiLU function defined previously is a special case of PSwish, when $\beta=1$. PSwish transitions from the identity function for $\beta=0$, to ReLU for $\beta \to \infty$. The curvature of PSwish increases as $\beta$ increases. Figure \ref{fig:parameteric_swish} shows the PSwish activation function for different values of $\beta$.

We show the results with the CIFAR-10 dataset, for final checkpoints for training and testing set in Table \ref{tab:parameteric_swish}. Interestingly, we observe that both the standard and robust generalization gap are extremely dependent on the choice of $\beta$. The robust generalization gap increases from $1.76$ to $38.57$  and the standard generalization gap increases from $1.82$ to $16.13$ for $\beta=0.5$ and $\beta=10$ respectively. We also observe that robust test accuracy for the final checkpoint increases from $45.24$ to $50.63$ for the same $\beta$ values. For larger values of $\beta$ i.e  $\beta \to \infty$, PSwish behaves like ReLU and standard and robust final test accuracy start decreasing.  The results are consistent with our previous experiments and show that the standard and robust generalization gap increases for activations with high curvature. Further using the early stopping checkpoint with the validation set, PSwish with $\beta = 10$ outperforms ReLU baseline by $0.33\%$ on robust accuracy and $1.24 \%$ on standard accuracy, highlighting that the choice of activation function can improve standard and robust performance for adversarially trained models. 

\begin{figure}
         \centering
         \includegraphics[width=0.4\textwidth]{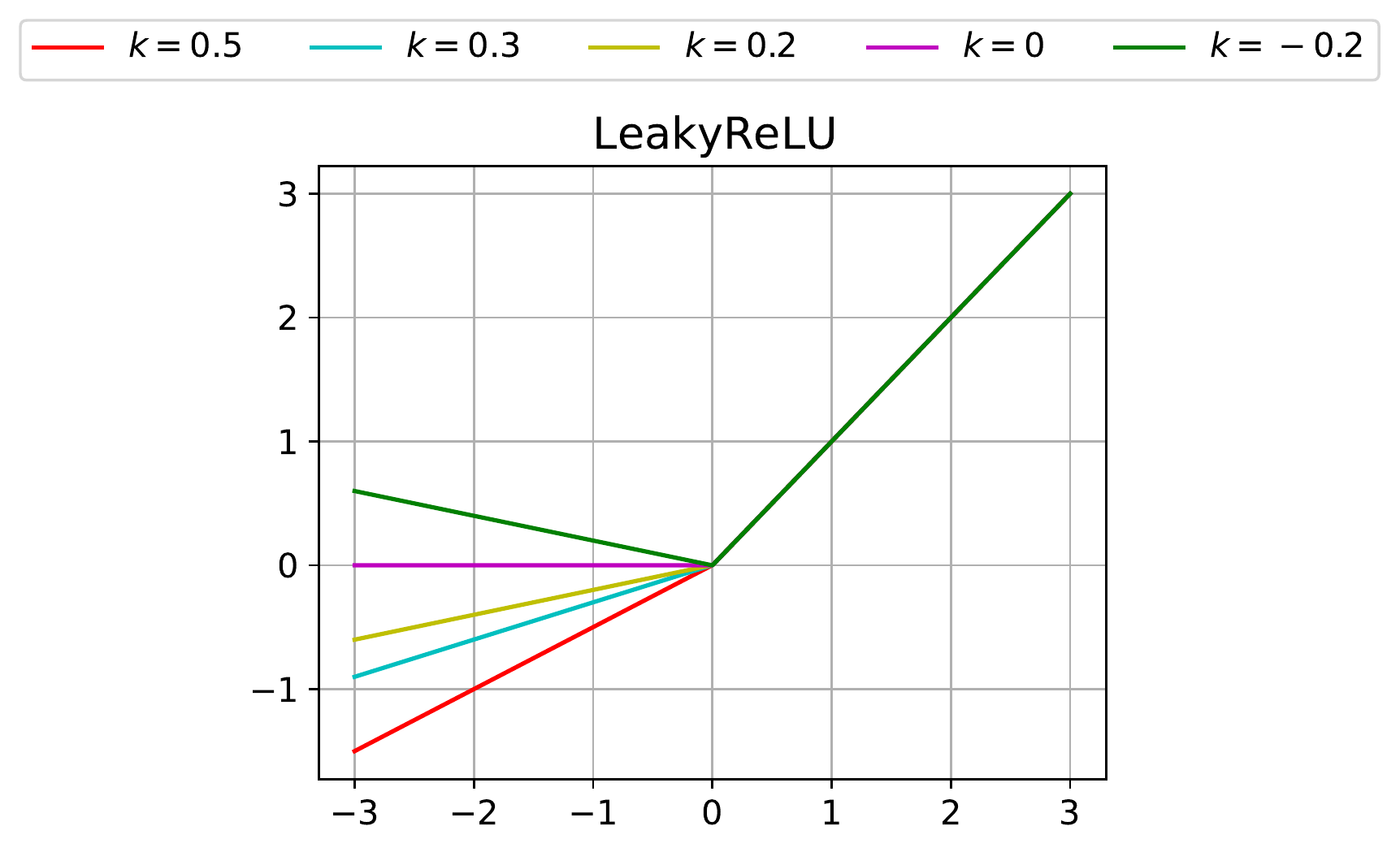}
         \caption{Visualization of LeakyReLU with different $k$ values.}
         \label{fig:leaky_relu}
\vspace{-2mm}
\end{figure}

\begin{table}[t]
    \centering
    \small 
    \begin{tabular}{c|c|c|c|c|c|c}
    k & \multicolumn{3}{|c}{Robust Accuracy} & \multicolumn{3}{|c}{Standard Accuracy} \\
     & Train &  Test & Diff. & Train & Test  & Diff.  \\
    \cline{1-7}
    0.5 & 52.74 & 48.71 & \textbf{4.03} & 82.99 & 79.56 & \textbf{3.43} \\
    0.3 & 63.06 & 49.62 & \textbf{13.44} & 92.00 & 83.56 & \textbf{8.44} \\
    0.2 & 69.64 & 49.34 &  \textbf{20.3} & 95.37 & 84.21 & \textbf{11.16}\\ 
    0  & 82.46 & 49.47 & \textbf{32.99} & 98.9 & 83.73 & \textbf{15.17} \\
    -0.2 & 85.89 & 48.16 & \textbf{37.73} & 99.47 & 83.01 & \textbf{16.46} \\
    \end{tabular}
    \caption{Performance of the LeakyReLU activation function with different slope values. The standard and robust generalization gap increases for slopes with larger  approximate curvature.}
    \label{tab:leaky_relu}
\vspace{-3mm}
\end{table}


\section{Does smoothness matter?}

 Xie \etal \cite{xie2020smooth} showed that using smooth activations, adversarial training can achieve better standard and robust accuracy on Imagenet \cite{imagenet_cvpr09}. They posit that using smooth activations can improve gradients, which can both strengthen the attacker and provide better gradient updates to weight parameters, thus achieving superior performance.

In contrast, we show that the relation of the generalization gap to activations can be observed for non-smooth activations as well. We use the non-smooth LeakyReLU activation function defined as follows:
\[
  \text{LeakyReLU}(k, x) =
  \begin{cases}
                                   x & \text{if $x \geq 0$} \\
                                   kx & \text{if $x < 0$} \\
  \end{cases}
\]
where $k$ is a hyper-parameter that can be tuned. The first derivative of LeakyReLU is given as:
\[
  \dfrac{d}{dx}\text{LeakyReLU}(k, x) =
  \begin{cases}
                                   1 & \text{if $x \geq 0$} \\
                                   k & \text{if $x < 0$} \\
  \end{cases}
\]
For non-smooth activations, curvature of the activation function however is not well defined. Therefore for LeakyReLU, we use the difference of slopes, i.e $|1-k|$ as the ``approximate'' curvature of the function. Hence, for $k \leq 1$ the approximate curvature decreases with increasing value of $k$ .
We use the same setup as in previous experiments and show the results for final training and test checkpoints in Table \ref{tab:leaky_relu} on CIFAR-10. We observe behavior similar to smooth activations for LeakyReLU. For $k=0.5$, the approximate curvature is low, and both robust and standard generalization gap, $4.03$ and $3.43$ respectively is much smaller than for $k=-0.2$, for which robust and standard generalization gap, $37.73$ and $16.46$ is large. We therefore hypothesize for non-smooth activations, the ``approximate" curvature of the activation function has impact on the generalization gap. 

\begin{figure}[t]
\begin{center}
\includegraphics[scale=0.45]{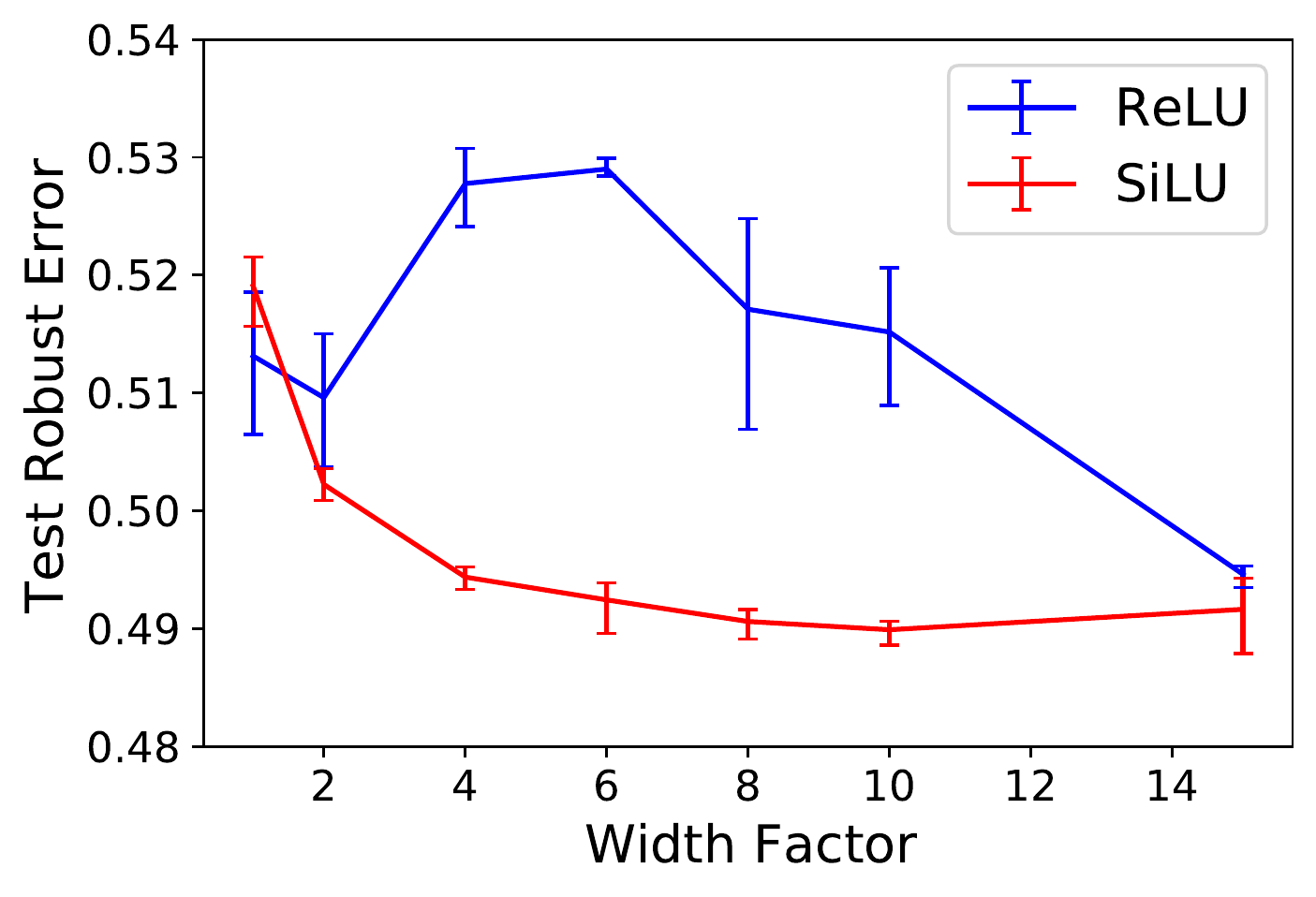}
\end{center}
\vspace{-1mm}
\caption{The generalization curves show the double descent phenomenon occurs for networks with ReLU activation but does not occur with SiLU activation. We use adversarially trained WideResnet models and the model complexity is controlled by the width of the architecture. Each data point shows the average for last 3 epochs. }
\label{fig:double_descent_curves}
\vspace{-3mm}
\end{figure}

\section{Double descent curves}

The standard bias-variance trade-off from classical machine learning theory fails to explain why deep networks generalize well especially when they have far more parameters than the samples they are trained on  \cite{zhang2021understanding}. Its now standard practice to use overparameterized models and allow models to train longer \cite{hoffer2017train} since test time performance typically improves for increased model complexity beyond the data interpolation point, a phenomenon known as \emph{double descent} \cite{belkin2019reconciling}. It was further shown that both training longer and increasing architecture size can be viewed as increase in model complexity and the double descent phenomenon is observed for both settings \cite{nakkiran2020deep}. The phenomenon of double descent generalization with increase in model width was also briefly noted for $l_2$ adversarially trained models \cite{nakkiran2020deep}.

Rice \etal \cite{rice2020overfitting} show that robust overfitting contradicts the double descent phenomenon observed with respect to training longer, since training longer harms test time performance. Although, they still observe the double descent phenomenon for ReLU networks with respect to the model size as shown in Fig. \ref{fig:double_descent_curves}. They therefore posit that, training longer and increasing model size have separate effects on robust generalization.  

 A recent work \cite{nakkiran2021optimal} suggests that the double descent phenomenon can be mitigated by optimal regularization. We explore whether activations with low curvature can mitigate double descent, by adversarially training Wide Resnets with different width factors. We show results for ReLU and SiLU activation functions in Figure \ref{fig:double_descent_curves}. Experiments with other activations could not be conducted due to the high expense of training Wide Resnets. We use the SiLU  activation function, because it has lowest curvature among all the activations considered. In Figure \ref{fig:double_descent_curves}, we show the results for ReLU and the SiLU activation function with a PGD-10 adversary.  While the double descent phenomenon is observed for ReLU activation, robust test performance continues to decrease for the SiLU activation function. Note that SiLU with \emph{width-factor 4 attains equivalent performance to ReLU with width-factor 15}.  None-the-less the final test error achieved by ReLU networks with large width factor is equivalent to the lowest test error achieved by SiLU networks with the same width. This suggest that low curvature activations may not be useful for models with large width. The results also indicate that use of activations with small curvature can act as a regularizer to mitigate the double descent phenomenon.  

\section{Conclusion}
In this work, we first use both theoretical and empirical approaches to show the impact of curvature of the activation function on robustness. We further show that this property of regularization further extends to non-smooth activations as well.
While results from Rice \etal show that classical regularization techniques are unable to prevent robust overfitting, our results show that activation functions with low curvature can largely mitigate that. Since robust overfitting is common in adversarial training, the properties of activation functions that we bring to light in this work can be useful for state of the art robust models. Finally our experiments also show that double descent, another phenomenon that has a significant impact on robust generalization, can be mitigated using activations with low curvature.

\section{Acknowledgments}

This project was supported in part by NSF CAREER AWARD 1942230, HR00112090132, HR001119S0026, NIST 60NANB20D134 and ONR GRANT13370299, Quantifying Ensemble Diversity for Robust Machine Learning (QED for RML) program from DARPA and the Guaranteeing AI Robustness Against Deception (GARD) program from DARPA. We are grateful to our colleagues Abhay Yadav, Songwei Ge, and Pedro Sandoval for their valuable inputs on the early draft of this manuscript.

{\small
\bibliographystyle{ieee_fullname}
\bibliography{main}
}

\end{document}